# Analyzing mixed construction and demolition waste in material recovery facilities: evolution, challenges, and applications of computer vision and deep learning

Adrian Langley[1,2]
Matthew Lonergan[2,3]
Tao Huang[1]
Mostafa Rahimi Azghadi[1]

[1] College of Science of Engineering, James Cook University, Townsville, Queensland, Australia
[2] Bingo Recycling, Sydney, New South Wales, Australia,
[3] Machine Vision AI, New South Wales, Australia

Correspondence
Mostafa Rahimi Azghadi, PhD, College of Science and Engineering, James Cook University, Townsville, QLD, Australia.
Email: Mostafa.rahimiazghadi@jcu.edu.au

Funding information
This research was supported by an Australian Research Training Program (HDR) Scholarship.

**Abstract**

Improving the automatic and timely recognition of construction and demolition waste composition is crucial for enhancing business returns, economic outcomes and sustainability. While deep learning models show promise in recognizing and classifying homogenous materials, the current literature lacks research assessing their performance for mixed, contaminated material in commercial material recycling facility settings. Despite the increasing numbers of deep learning models and datasets generated in this area, the sub-domain of deep learning analysis of construction and demolition waste piles remains underexplored. To address this gap, recent deep learning algorithms and techniques were explored. This review examines the progression in datasets, sensors and the evolution from object detection towards real-time segmentation models. It also synthesizes research from the past five years on deep learning for construction and demolition waste management, highlighting recent advancements while acknowledging limitations that hinder widespread commercial adoption. The analysis underscores the critical requirement for diverse and high-fidelity datasets, advanced sensor technologies, and robust algorithmic frameworks to facilitate the effective integration of deep learning methodologies into construction and demolition waste management systems. This integration is envisioned to contribute significantly towards the advancement of a more sustainable and circular economic model.

## 1. Introduction

The amount of construction and demolition waste (CDW) sent to landfills or material recycling facilities (MRFs) is a dynamic issue influenced by a combination of economic, environmental and regulatory factors. The global trend of increasing CDW poses significant challenges for existing MRFs in maintaining efficient sorting, processing and recovery rates. In Australia, 29 million tons (38% of total waste) of CDW were generated between 2020-21, representing a 24% increase over the previous 4-year period (Pickin et al., 2022). The sheer volume and diversity of waste materials is a challenge for the scalability of recycling systems and the efficiency of processing materials. Significant investment in MRF infrastructure, equipment, advanced sorting technologies and or sensors or regulatory changes may be required to meet this demand (Ali and Courtenay, 2014).

Identifying the types of waste in the waste stream has been a significant obstacle in improving the recycling rate (proportion of materials recycled or recovered). Current research in sorting and segregation largely focuses on advances in optical sensor technology (image, spectroscopic, spectral, etc.), artificial intelligence (AI) or a combination of both (Pučnik et al., 2024). Deep learning (DL) models for recognition and classification of CDW show promising results with high levels of accuracy; however, these are usually trained on datasets without complex backgrounds or heterogeneity of source material (Shahab et al., 2022). Challenges to successful implementation for commercial use include varying accuracy, lack of high-quality datasets for training and dynamic nature of construction sites (Majchrowska et al., 2022).

Intelligent detection and or segmentation using optical sensors or DL of waste in highly cluttered collections remains only a small fraction of the CDW research in the past decade; even fewer studies have examined these in actual commercial MRFs (Prasad and Arashpour, 2024). Adding to the complexity is the diversity of the waste streams, which may also include a wide range of contaminants, making generalization across different sorting scenarios or locations difficult. Real-world sorting or transfer centers can be harsh, with physical and environmental factors influencing accuracy. Sensor degradation can occur due to high dust, debris and moisture levels, lighting variability and interference from other systems and equipment, for example, vibration, affecting performance (Kroell et al., 2022).

In Australia, most companies use skip bins to collect CDW material. A top-down or surface assessment can be problematic (Driouache et al., 2024). Bins may have lighter material, such as pallets, stacked on top of heavier mixed waste (Davis et al., 2021). Relying on surface features alone may also introduce bias; contractors may load trucks in a particular way depending on accessibility to the waste or stage of demolition (Chen et al., 2021). Figure 1(a) shows a skip bin being delivered to an MRF, with its contents ready to be tipped. The waste pile is then spread using an excavator (Figure 1(b)) before being loaded onto a conveyor belt by a hydraulic grab. Figure 2 demonstrates the subsequent processes of that waste in the MRF. The colored box is the focus of this review. This step often requires manual inspection to check for prohibited or bulky items that pose a safety risk to personnel or equipment downstream. Analysis of this pile not only informs operational decisions but also highlights the critical need for advanced technologies, such as deep learning, to enhance the accuracy and efficiency of material identification and processing. Given the importance of these advancements, a comprehensive review is necessary to understand the current state of the field and identify pathways for future progress.

This review aims to systematically examine and synthesize existing literature on DL applications for CDW management, with a particular focus on object detection and segmentation for real-time implementation in MRFs. Current research mainly focuses on discrete, homogenous waste piles or mixed waste streams on conveyor belts. In contrast, this paper reviews advancements in DL technology and examines case studies of DL-assisted CDW recycling, highlighting successes and persistent challenges limiting commercial adoption. This review underscores the need for future research to harness DL methods, driving significant improvements in CDW recycling processes and supporting the industry's shift towards a circular economy.

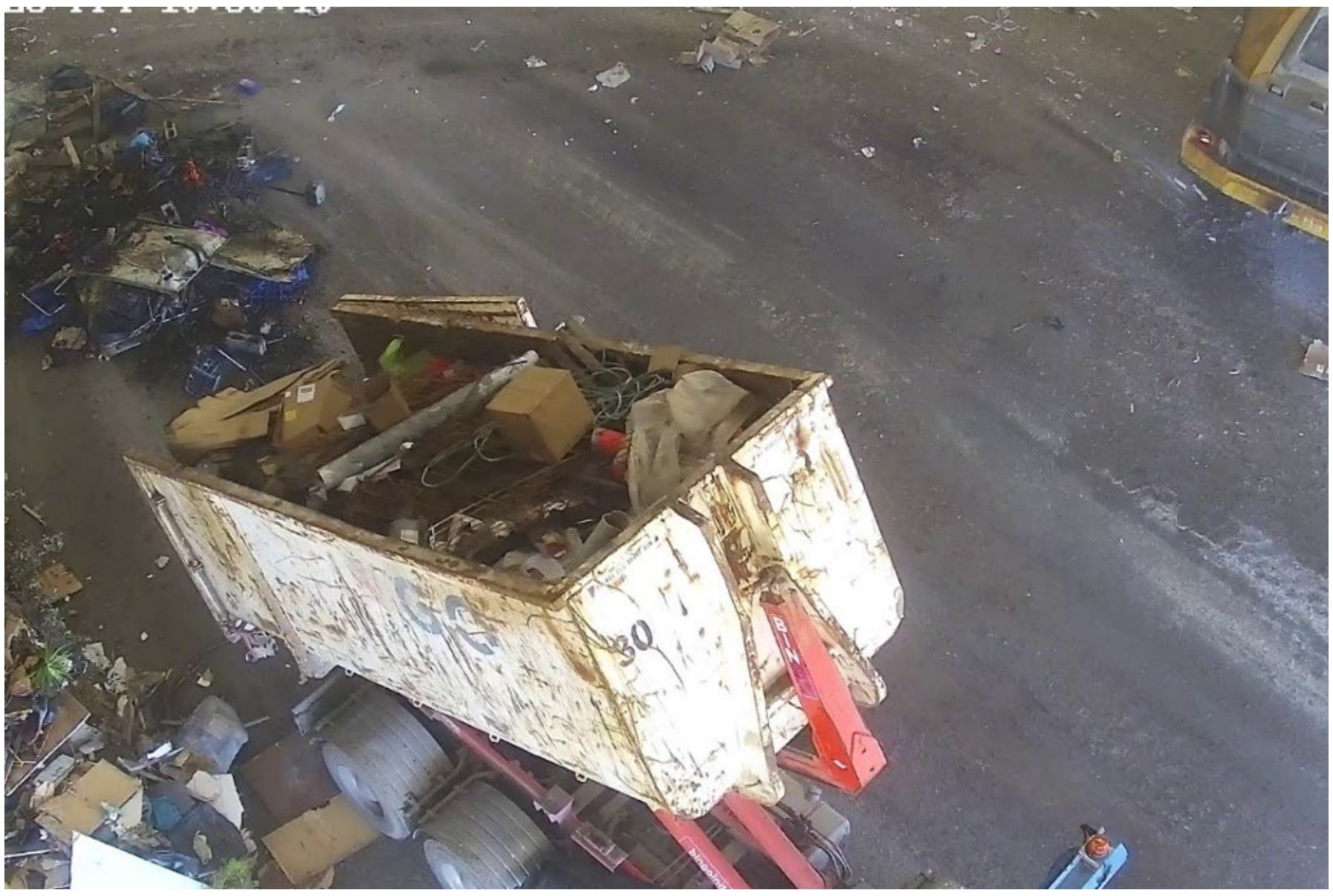

**(a)**

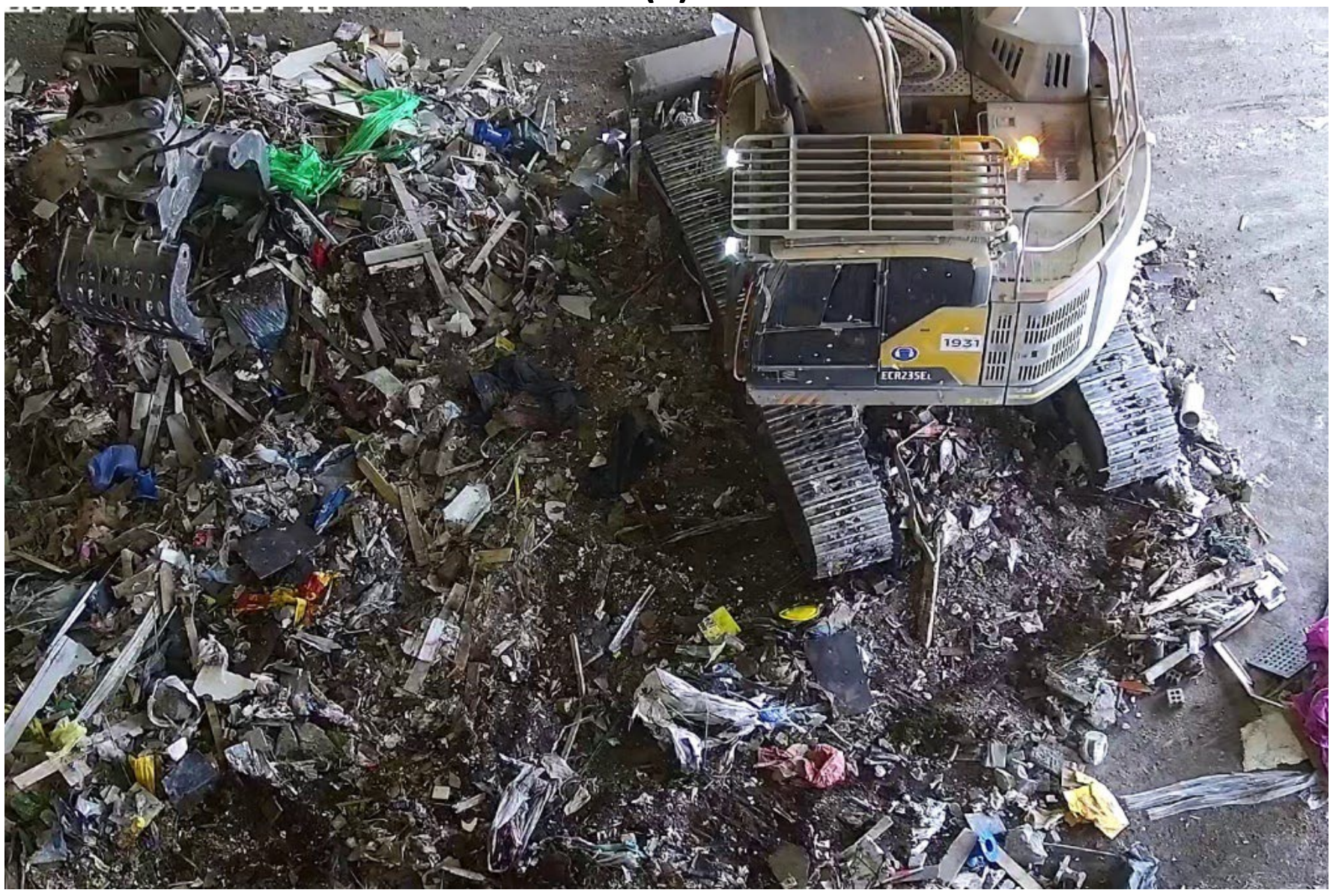

**(b)**

**Figure 1.** Real world example of CDW waste. (a) an example of a typical mixed CDW-containing skip bin. (b) Pile waste being spread and examined with an excavator.

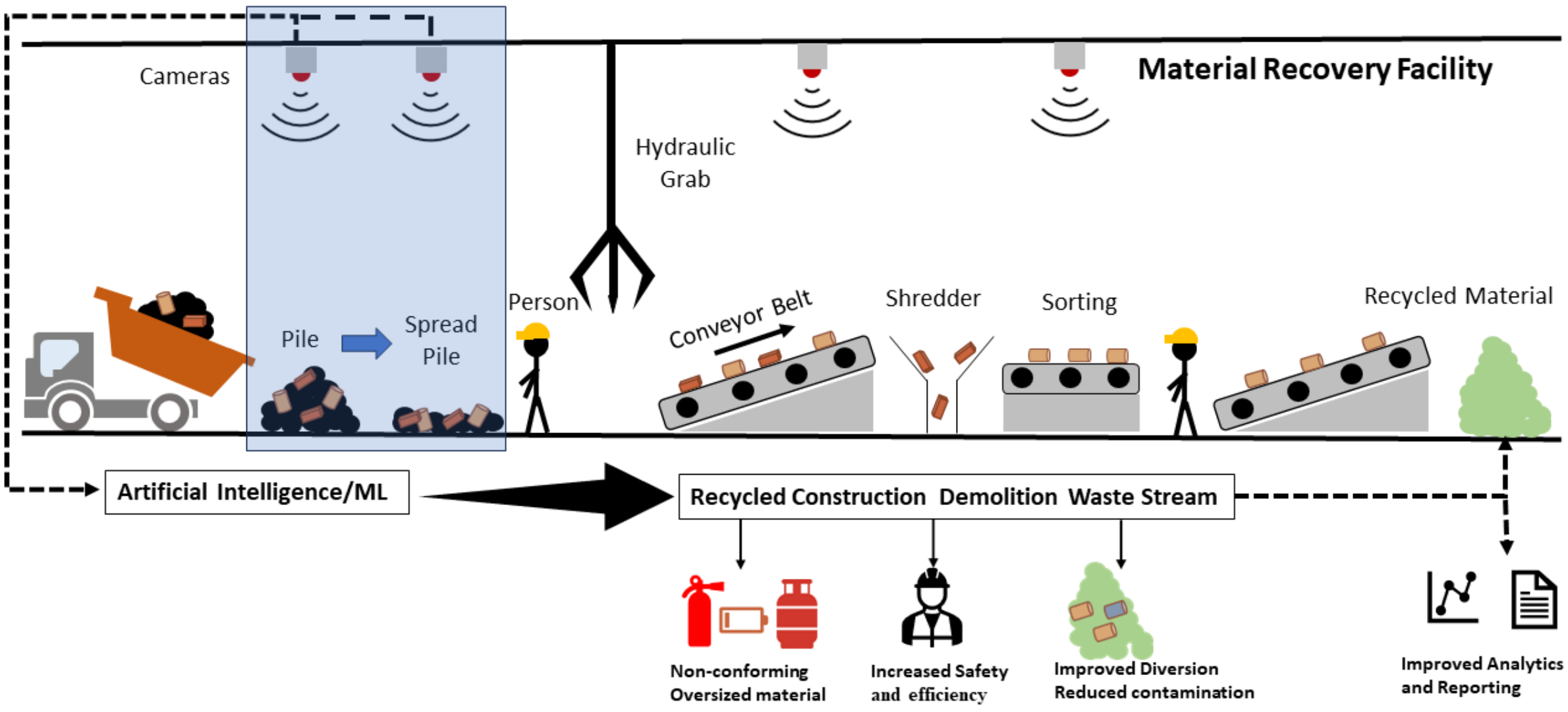

**Figure 2.** Overview of the CDW recycling stream with various processing stages, including sorting and shredding. The colored box is the focus of this review.

## 2. Research methodology

This paper reviews DL technologies in CDW management to better understand the progress towards real-time composition analysis and object detection/segmentation. This literature review relied heavily on the quality of its initial search terms. To identify the most relevant keywords, a preliminary review was conducted on construction and demolition waste recycling, artificial intelligence, circular economy, computer vision, sensors, material recycling facilities, and deep learning. Figure 3 illustrates the flowchart of the paper selection process.

Scopus was selected as the primary database due to its extensive coverage of peer-reviewed journals in relevant fields and its frequent use in construction engineering research (Pal and Hsieh, 2021). The search encompassed article titles, abstracts, and keywords within the specified timeframe (2004 - 2025). The same search criteria were also used for Web of Science (WoS) core collection. Google Scholar was also searched to identify reports, technical papers, and other materials that may not be indexed in traditional databases like Scopus or Web of Science. The bibliograph of these reports was examined to find relevant references. Although only peer reviewed sources were analyzed, this was crucial for accessing potentially valuable research in this rapidly evolving area.

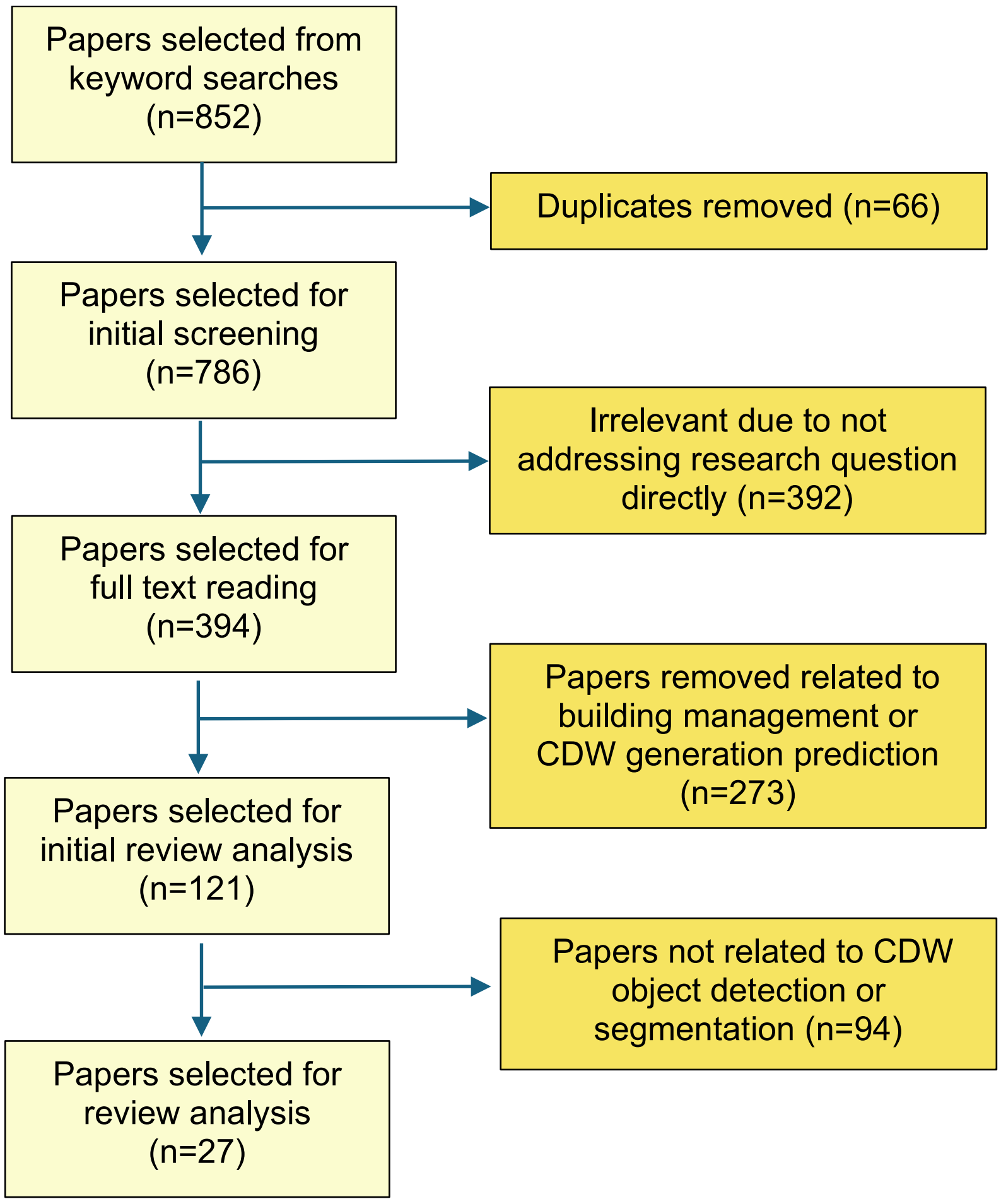

**Figure 3**. Flowchart of paper selection process.

The search field in Scopus was set as "Article title, Abstract, Keywords". The first set of searches consisted of "construction AND demolition AND waste (management OR recycling)" and deep learning OR CNN OR transformer OR object detection OR segmentation OR conv* OR artificial intelligence. The second keyword set consisted of "construction waste AND learning" AND sensor OR dataset. The keywords within each set were combined with 'AND' or "OR" operators with '*' symbol used for related word variants. The third key word search consisted of "material recovery facility" AND recycling AND deep learning OR sensors. The same searches were conducted in WoS using 'All Fields' and Google Scholar.

The initial search retrieved 692 papers from Scopus and 94 from WoS. The lists were integrated by removing duplicate entries, conditional formatting applied to the 'Article Title' and a manual review of each abstract for relevance performed. Document types such as articles, conference papers, conference reviews, reviews, and book chapters written in English were used for this review. The integration process retained 121 papers for analysis, excluding those on solid or municipal waste and deep learning studies unrelated to CDW recycling or management. These 121 papers served as a foundation for the review helping to understand the landscape of CDW research; however, given the focus on deep learning, only research from the past five years on segmentation and object detection was considered, resulting in 27 papers listed in Table 1.

Table 1 presents the information retrieved from the relevant papers organized for clarity and comparison. The fields include type of CDW materials, sensors used to collect the data, method or models investigated, research setting, image resolution (if reported), and dataset size. Each publication year has been noted to highlight the increasing volume of publications in recent years. Analysis of Table 1 highlights the diverse CDW material datasets, sensors, and deep learning models used for object detection and segmentation. The following sections examine each of these topics, exploring their opportunities and challenges.

## 3. Challenges and opportunities in developing CDW in-the-wild datasets

DL algorithms require large, varied datasets to extract high-level, complex abstractions; no single algorithm can guarantee the same results across all datasets. The growth of datasets designed for CDW recycling has evolved over the years since the early benchmark for waste analysis were proposed in 2016 (Yang and Thung, 2016). Early approaches dealt with single-stream datasets, which were too simple for commercial applications and did not reflect the variety of CDW materials. Table 1 highlights not only the evolution and complexity of these datasets but also the challenges of comparing research results. There exists a wide range of dataset diversity including vary image quality and resolution, location settings and types of material included (Lu and Chen, 2022).

MRF specific datasets remain a small part of the current research domain and have only started to appear in recent years. The Construction and Demolition Waste Object Detection (CODD) (Demetriou, 2022) and ZeroWaste Datasets (Bashkirova et al., 2021) attempt to capture the composition complexities of CDW through high levels of clutter and variation. However, these may still fail to provide realistic performance of model assessment in an MRF due to lack of variation in recyclable classes. These datasets may have limited utility in assessing CDW in mixed piles (Prasad and Arashpour, 2024). The annotation and expertise-related expenses associated with developing fully supervised DL algorithms can be prohibitive (Munappy et al., 2022). To fill this gap, techniques such as data augmentation and synthetic data creation may be necessary to develop commercially viable algorithms.

### *3.1. Data augmentation*

Data augmentation is a technique to artificially expand a training dataset by creating additional, slightly modified versions of existing data. The process involves applying a set of transformations or manipulations of the original data that preserves its label or class while creating new data points that are similar but not identical to the original. Image augmentation can include transformations such as rotation, flips, color jittering, geometric transformations, cropping and changes in brightness or contrast. The technique helps deep learning by increasing the data to train models effectively, reducing overfitting and improving robustness through training a model on a more extensive and diverse dataset (Shorten and Khoshgoftaar, 2019). In one study, increasing the amount of CDW data through augmentation using transfer learning increased the mAP by 16% (Na et al., 2022).

### *3.2 Synthetic CDW data*

Synthetic data refers to artificially generated data that mimics the statistical properties of a real-world dataset. It is often used when real-world data is challenging to obtain, expensive, or private and sensitive (Nikolenko, 2021). It can also be used to augment existing datasets or to create simulations for testing and training purposes. Privacy is less of a consideration for CDW datasets than in other fields, such as medicine; however, these may be commercially sensitive (Rajotte et al., 2022).

Software, including Blender 3D and Unity, have been used to create 3D synthetic data for DL models for PET recycling and waste sorting (Feščenko et al., 2023). A gaming engine (Unreal Engine 4) was used to generate synthetic data with similar distance, orientation, camera rotation, texture and light source to balance classes for image detection using CNN-based YOLOv5. The model utilized purely synthetic data to identify several categories of objects, including pallets and crates, on a portable Jetson Nano single-board computer with a RealSense Depth Camera D435i (Rasmussen et al., 2022). Although this work shows promising results using synthetic data in the CDW visual analysis, some limitations should be considered. Models trained purely on synthetic data may not generalize well if the data does not capture the complexity and variability of real-world conditions; however, fine-tuning models trained on large synthetic datasets on a few real images may increase real-world performance (Baaz et al., 2022).

### *3.3 Large language models (LLM) and Segment Anything (SAM)*

Recently, the release of several computer vision foundation models, such as Segment Anything (SAM) (Kirillov et al., 2023), SAM-2 (Nikhila et al., 2024), DINOv2 (Oquab et al., 2023), and CLIP (Radford et al., 2021), have greatly stimulated research in the CV community and have tremendous implications for CDW modelling. Although some examples focus on text and language, these models can also assist segmentation through zero-shot recognition or generate images through prompts such as speech-to-text-to-image/video. Given their strong generalization capabilities, LLMs (Frei and Kramer, 2023) may offer significant advantages for developing or utilizing assisted segmentation techniques to efficiently create datasets for contaminated CDW.

Meta's SAM is a foundation CV model for segmentation and has been applied to numerous domains, from medical (Zhang et al., 2023) to the construction industry (Ahmadi et al., 2023). The potential to automate labelling CDW datasets makes model development easier and may improve the time and accuracy of resultant datasets. SAM's zero-shot abilities across various tasks are impressive as an out-of-the-box tool, but CDW introduces nuances not seen in everyday images. It is currently unknown if domain-specific fine-tuning, incorporating expert annotations, domain-adaptive techniques or modification to the algorithm structure, through methods such as patch inference, is needed to improve for specific circumstances or datasets (Xie et al., 2024).

To address this question, Panizza et al. studied five classes of material (bricks, concrete, PVC pipes, plastic wire and rebars) obtained from the Site Object Detection Dataset (SODA), Google and synthetic model generators (Shutterstock, 2024) and Sketchfab (Sketchfab Inc, 2024). The dataset consisted of 1276 images, divided into training (80%) and testing (20%) fractions. The images were labelled using LabelMe (Wada, 2018) and SAM. The authors concluded that SAM improved the ease of labelling but with a maximal IoU loss ranging from 6.6% to 28.35%, depending upon the material (Panizza et al., 2024). This is consistent with similar studies from other domains, for example, medicine (Ferreira and Arnaout, 2023).

LLMs came into prominence in 2018 with subsequent iterations, including Open AI ChatGPT and Google's Gemini (Schalkwyk et al., 2023). These perform well on diverse tasks in domains including business and education; however, little work has been done in the field of CDW recycling (Saka et al., 2024). Multi-modal LLMs can also generate images; text prompting can allow precise control over image content, allowing modification or customization of backgrounds, styles and content. This may enable the rapid generation of larger CDW datasets whilst reducing costs and allowing the visualization of desired mixtures or components. While improving dataset creation times, these synthetic images might still have subtle artifacts or unrealistic elements that source content experts can identify (Cao et al., 2024). Figure 4(a) illustrates an image generated with Google's Gemini. The text prompt was "show an image, using an overhead top-down view, of a waste tipped from a skip bin onto the floor, which has been spread out, with construction waste consisting of 50% concrete and 50% wood". Figure 4(b) demonstrates the same text prompt using DeepSeek (Lu et al., 2024). Other settings included natural lighting, realistic art style and neutral mood.

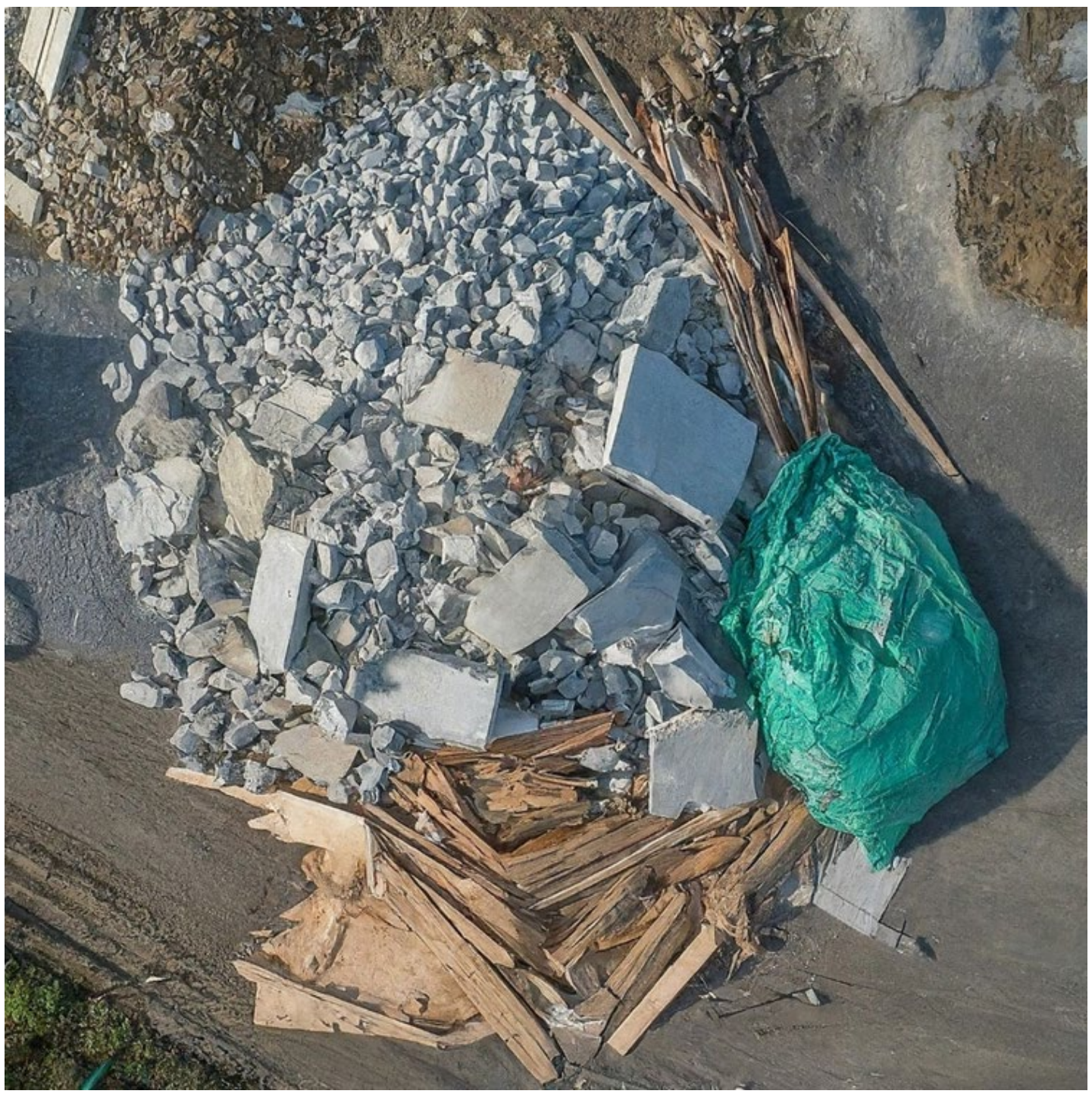

**(a)**

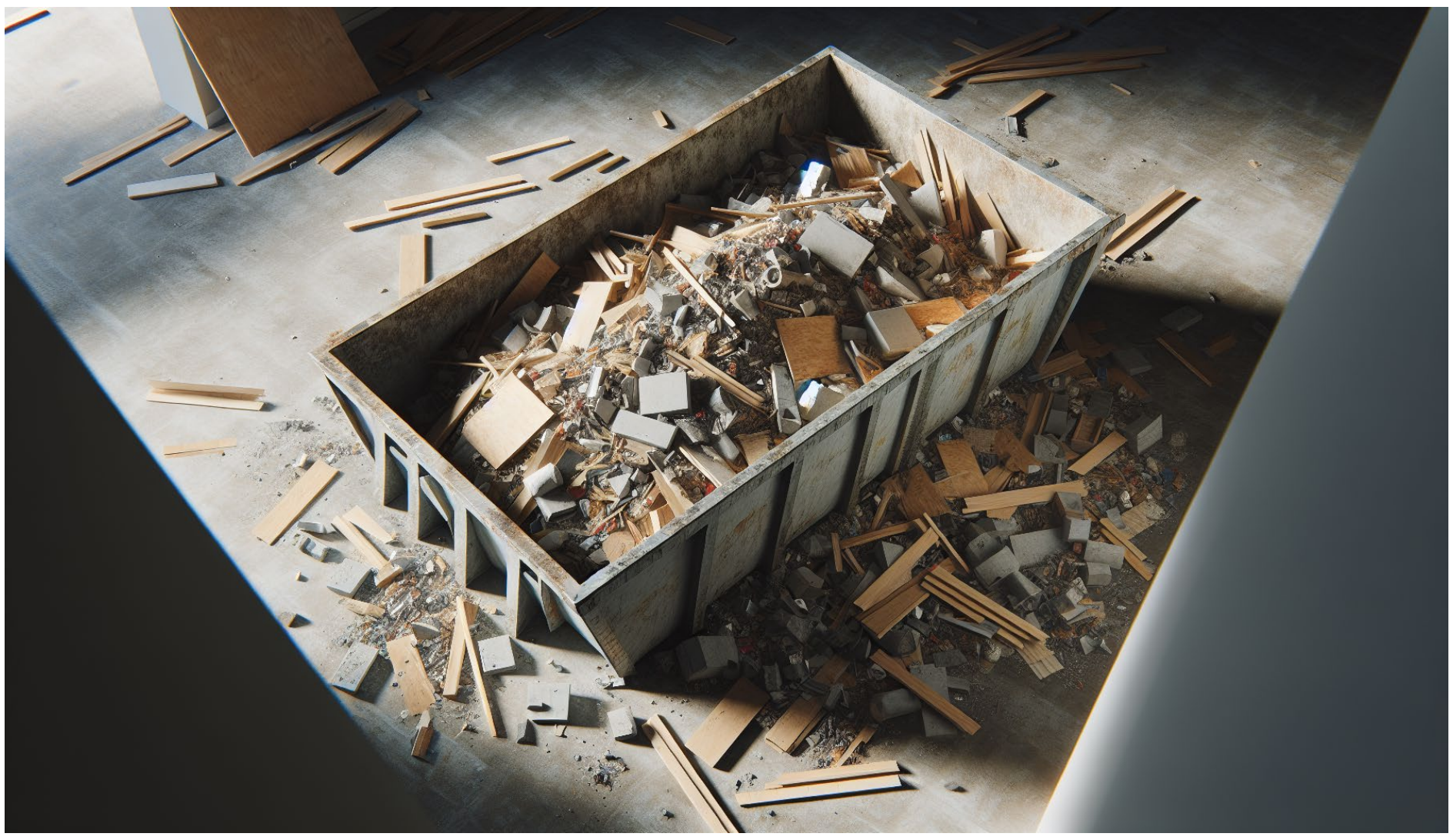

(b)

**Figure 4.** Synthetic images generated by (a) Google's Gemini, and (b) DeepSeek, asked to create an image of CDW waste pile containing concrete and wood.

**4. Sensors for CDW analysis**

The use of different sensors in DL systems for CDW recycling has progressed notably. Initially, the emphasis was on traditional imaging methods, where approaches centered around basic visual checks for recycling processes. However, as technology evolved, more advanced sensors have been utilized to boost the accuracy of material identification in MRFs (Chen, L. et al., 2024). As early as 2002, edge detector algorithms were used to detect pixels that belong to the target object (concrete, steel, timber, aluminum) by comparing the RGB value of the pixel with the RGB range of the material from which the target is made. Once the edges have been detected, the pixels were grouped inside them to form an object (Abeid Neto et al., 2002).

In 2010, the detection of concrete in construction site images relied upon predefined color/texture value ranges for material recognition with varying thresholds. With manual color and texture features, automatic concrete detection correctly identified concrete using neural networks and SVMs with an average precision and recall of around 80%. Missing classification of small regions (< 200 pixels) occurred as part of the pre-processing and image segmentation process (Zhu and Brilakis, 2010); however, small object identification remains a problem today with even more complex and sophisticated object detection algorithms (Chen, X. et al., 2024).

By 2014, accuracy of greater than 97.1% were obtained using the SVM classifier on images from the Construction Materials Library (20 typical construction classes) for high-quality 200 x 200-pixel color images. It averaged above 90% for small 30x30 pixels and 92% for highly compressed, low-quality images under real-world conditions, improving upon previous results of 70% (Dimitrov and Golparvar-Fard, 2014). The relationship between sensor technologies and DL suggested the potential of integrating real-time data gathering with DL models to enhance sorting and recovery processes in CDW recycling. This trend reflects an increasing acknowledgment of the significance of advanced sensor technologies in boosting deep learning capabilities, ultimately leading to better material recovery solutions (Choi et al., 2023).

*4.1 RGB-depth cameras*

RGB cameras are widely used visual sensors in many DL applications; in many scenarios, providing enough information for a DL model to perform a recognition task properly (Qiao et al., 2024). In MRFs, dust may cover object surfaces, reducing camera recognition accuracy. Adding in depth information, which is not easily affected by dust, color, or lighting changes, may improve detection rates. However, in one model using laser line scanning for depth information, classification accuracy was only modestly increased by 1.92 ~ 2.27% (Li et al., 2022).

A further problem for RGB cameras in CDW identification is heterogeneous sample composition within a single category. Wood, for example, can be derived from natural or engineered sources. Phenotypically similar materials such as aerated, lightweight and porous concrete may have different optical properties. Similar-looking materials, such as natural aggregates, have higher water adsorption with a lower grain strength than concrete (Anding et al., 2011). Therefore, these similarly looking materials cannot be easily identified by visible spectrum (i.e., RGB) necessitating other sensor types.

### *4.2 Near-infrared*

Near-infrared (NIR) spectroscopy, encompassing wavelengths from 700 to 2500 nanometers, elucidates the chemical composition of a sample by analyzing the characteristic vibrational transitions induced by the absorption of infrared radiation (Emsley et al., 2022). NIR may improve plastic identification compared to standard image identification but may have some shortcomings, particularly in identifying black materials, as these are less distinguishable. Adding NIR spectral recognition increases the cost and complexity of systems. In the context of CDW, many inorganic substances in construction waste do not contain distinct functional groups absorption spectra, making NIR spectroscopy difficult without initial pre-processing of the spectral curve (Xiao, W. et al., 2019).

### *4.3 Hyperspectral imaging*

Hyperspectral imaging (HSI) is an emerging rapid and non-destructive technology that may have promising applications for the identification of CDW. HSI offers a potential advantage over NIR by integrating spectroscopic and visible imaging capabilities within a single system. This unique approach enables simultaneous acquisition of both spectral and spatial information (Tahmasbian et al., 2021). It has been used to separate different types of plastics in municipal solid wastes and to recover and recycle concrete, mortar aggregates, bricks, tiles and wood (Castro-Díaz et al., 2023). HIS images are costly due to the complicated apparatus required to acquire a wide continuous spectrum. Still, they are also more robust to solar reflections, a significant concern for RGB cameras. Reconstruction of HIS from single RGB images using convolutional neural networks (CNNs) offers a portable, low-cost alternative (Gao et al., 2021), which can be adopted in the CDW domain but requires significant research and development.

### *4.4 Multi-sensor fusion*

The composition of CDW may be complex, resulting in equally complex spectral features obtained from NIR or HIS techniques. In such cases, multi-sensor fusion and the internet of things (IoT) may provide agile solutions for classification and real-time monitoring of municipal solid waste (MSW) (Mookkaiah et al., 2022). Such techniques may also be introduced into CDW recognition. Emerging 5G and 6G networks enable IoT infrastructure that will allow higher resolution images and integration of wireless sensor networks to track the CDW at the source and direct it to the appropriate MRF (Jagan and Jayarin, 2022).

*4.5. Other sensors*

Other sensors that can be fused for CDW recognition include infrared, ultrasonic, line and laser scanning, weight and chemicals. Lidar sensor fusion leverages detailed 3D information, whereas ultrasonic can provide high precision distance measurement capability to improve waste sorting and efficiency (Aliew, 2022). Combining multiple sensors can provide a more comprehensive understanding of waste streams but at increased expense to integrate into existing infrastructure and may be subject to calibration, data synchronization and complexity or processing issues. Algorithms using sensor fusion can require significant technical expertise. Sound recognition, metal detection and weight were used to detect glass and metal in trash bases based on spectrograms in highly controlled circumstances with an accuracy of 98% (Funch et al., 2021); however, this is impractical for a large MRF.

## 5. Deep learning for visual CDW analysis

The application of DL technologies for CDW analysis in recycling facilities has evolved significantly over the past decade. Initial studies focused on traditional CV techniques, which often struggled with the complexities and variabilities inherent to CDW streams (Lin, K. et al., 2022). Early work highlighted the need for enhanced detection methodologies to manage the processing of diverse materials effectively (Lopes et al., 2024). As the field matured, researchers began to explore machine learning approaches that could automate the classification and sorting of CDW with improved accuracy, laying the groundwork for subsequent innovations in deep learning (Wang et al., 2020).

### 5.1. Object detection for visual CDW analysis

In a short space of time, driven by the increasing speed of object detection algorithms and the need for real-time sorting, classification has given way to object detection and segmentation algorithms (Diwan et al., 2023). MaskRCNN has been used with robot models for CDW detection. In one study, using an RGB-Depth camera, a real-time sorting robot was able to analyze CDW on a conveyor belt with a speed of 0.25 meters per second and inference time not exceeding 1920 milliseconds (Li et al., 2022). Despite its accuracy, MaskRNN's two stage architecture makes it less suitable for real-time applications compared with single stage algorithms (YOLO, SDD). Classifying materials from 15 to 25 meters was also challenging for RGB cameras; this poses a problem for cameras placed too far from the load drop off zone, or if those zones vary on the MRF floor (Wolff et al., 2021).

In the last few years, the research landscape has expanded beyond object detection to include comprehensive analyses of CDW composition. Demetriou *et al.* assessed 18 models, both single-stage (SSD, YOLO) and two-stage (Faster-RCNN) detector architectures coupled with various backbones feature extractors (ResNet, MobileNetV2, EfficientDet) on 6600 CDW samples belonging to brick, concrete and tile under working conditions with normal and heavily stacked adhered samples. YOLOv7 attained the best accuracy (mAP50:95 ~ 70%) at the highest inference speed of less than 30ms (Demetriou et al., 2023).

### 5.2. Object segmentation for visual CDW analysis

Compared to object detection, segmentation can provide more precise information about mixed objects or regions within the image. Semantic segmentation performs classification at the pixel level, which may provide more granularity on waste composition compared with classification/object detection models. It is currently one of the most popular research fields in CV. It underlies technology in autonomous vehicles, medicine, image search engines, industrial inspection and augmented reality (Yu et al., 2023).

YOLACT is a real-time instance segmentation algorithm that extends the YOLO framework by predicting object masks alongside bounding boxes and class labels. It has shown promising results in some studies, but its performance may be limited by the size and augmentation of the training dataset (Ko et al., 2022). Vision Transformers, a relatively new class of models, have demonstrated strong performance in various image recognition tasks, including object detection and semantic segmentation, and are considered a promising avenue for future research in construction and CDW applications (Dong et al., 2022).

The current leading segmentation algorithm for 'real-time' analysis of highly cluttered CDW, YOLOv9e-seg, achieves a mean average precision (mAP50:95) of 49.92 while processing at 125 frames per second. However, the mAP for small recyclable objects (<322 pixels) is significantly lower compared to medium (322–962 pixels) or large (>962 pixels) materials. The authors revealed that the examined models exhibited a bias towards objects situated within less visually complex environments. This preference could be attributed to the enhanced ease of object identification and segmentation within such contexts. The implementation of a patch-based inference strategy mitigated the detrimental impact of visual clutter on object detection performance, resulting in a mAP of 56.34. Notably, this improvement in detection accuracy was achieved without a significant compromise in classification accuracy or inference speed (Prasad and Arashpour, 2024).

## 6. Challenges and opportunities for DL in commercial CDW MRFs

### *6.1. Dataset challenge*

A look at existing literature, as outlined in Table 1, shows numerous themes tied to the challenges of CDW dataset generation. Generating standardized datasets for this industry is difficult and expensive; the makeup of waste shifts over time and varies by location, indicating the need for flexible and thorough datasets that can accurately capture these changes. The challenges of annotating these datasets highlight not only the technical hurdles in identifying and categorizing materials but also the demanding and sometimes subjective nature of the labeling process (Demrozi et al., 2023). Despite advancements in automated data collection methods like image recognition and sensor fusion, there is still a notable lack of standardizing datasets for deep learning in CDW waste recycling.

The lack of a common standard results in datasets being developed inconsistently, which can negatively impact the performance and adaptability of deep learning models in various recycling settings (Liang et al., 2020). Moreover, existing research often fails to address the differences found in real-world scenarios, where environmental factors and operational conditions can affect the accuracy and usefulness of the generated datasets. These issues stress the urgent need for more research to create best practices for dataset production, focusing on data quality, diversity and integration of both multi-modal and emerging sensor technologies with real-world recycling operations. Increased collaboration among academics, industry players, and policymakers could support the creation of standardized data collection protocols and promote the sharing of datasets across research and practical environments.

*6.2. Generative AI and domain specific prompt engineering opportunities*

Annotating complex CDW images is challenging due to the presence of small, innumerable, and difficult-to-identify materials. Stable Diffusion offers solutions by generating synthetic CDW images with controlled composition, augmenting existing datasets with label-preserving techniques, and even generating initial labels for human refinement (Valvano et al., 2024). Moreover, the rise of multi-modal AI (Barua et al., 2023), including the potential for CDW-specific models, combined with the increasing integration of IoT devices, promises to revolutionize CDW management by providing more comprehensive and accurate insights into recovery processes.

The advent of prompt engineering enables the use of pre-trained models for tasks by using customized prompts, which helps the models understand context better without needing a large amount of labeled data or complete retraining. Combining techniques like the YOLO system with infrared imaging shows how prompt engineering can improve detection accuracy in certain conditions by using different data enhancement methods (Yang et al., 2024). Overall, these strategies not only make the annotation process more efficient and cheaper but also set the stage for scalable solutions in many deep learning applications.

*6.3. Business case challenges and opportunities*

A business case supporting the purchase, development, installation, maintenance and use of a DL system should include an analysis of the expected costs and benefits. AI projects can be expensive and may not provide an immediate return on investment (ROI). The business case should start with a clearly defined problem that the model will solve and how it will improve efficiency, reduce costs, or open new opportunities. It should also quantify the problem, providing data demonstrating the extent and impact of the problem e.g., losses due to errors, time wasted on manual tasks, potential gains from new services or efficiency in current ones. The proposed solution should also outline the model functionality (classification, detection and segmentation) and the hardware or software services required (Enholm et al., 2021).

There are some spectacular examples of AI implementation failure (Olavsrud, 2024). Hence, knowing what an AI investment is worth and how to measure that value is a prerequisite for intelligent decision-making. An estimated 87% of data science projects fail to make it into production (VentureBeat, 2019). Simply investing money and expecting a high-tech solution at the end of the project does not happen without the proper leadership support and conditions for success. The costs for development can be considerable and vary significantly; small to medium-scale projects can cost from $10,000 to $500,000 (CHI Software, 2024). Development considerations include data collection and annotation, model development (engineers, data scientists), infrastructure, and services. Deployment costs may include integrating models into existing systems and other expenses, for example cloud computing. Maintenance considerations include model updates, retraining, monitoring performance drift, ongoing technical support, and hardware replacement (edge devices, routers, power supplies) (Smith, 2023).

Several frameworks (Bevilacqua et al., 2023) have been published to help estimate ROI, but what seems clear is that the productivity dividend of AI does not materialize immediately (Pandey et al., 2021). Specifically, for CDW processing, tangible benefits include reducing labor costs through task automation and reducing errors, increasing revenue through less plant downtime, speeding up processing times, and producing a cleaner product for marketing. Intangible benefits include improved safety or compliance, potential competitive advantage in the CDW recycling market and enhanced customer experience by providing more detailed and accurate information for billing and reporting purposes.

## 7. Discussion and Conclusion

The study of CV and DL technologies in mixed CDW analysis at MRFs shows new innovations but also points out major challenges that limit their effective use. Key results from this review show that while DL models, especially convolutional neural networks, have improved material classification accuracy, the diverse nature of mixed CDW waste creates serious problems. Differences in material types and insufficient training datasets reduce the effectiveness and widespread use of these technologies in practical commercial situations such as MRFs. More large volume, high-quality, and varied datasets that represent the complexities of mixed waste streams to enable strong algorithm training are needed.

Reinforcing the main point, it is evident from literature that combining computer vision, sensors, and DL in CDW analysis is a critical area in waste management technology. Additionally, it highlights the importance of interdisciplinary approaches that mix engineering, environmental science, and waste management to develop better and more sustainable solutions. Developing an international collaborative data repository for CDW images may help to address issues around dataset standardization, sensors modality and model performance (Kras et al., 2020), if commercial barriers can be overcome. The environmental benefits of recycling efforts can also be better assessed using the database. By studying trends and patterns in CDW disposal, regions could pinpoint their most pressing waste streams and customize strategies accordingly. For instance, areas with high amounts of materials, like concrete or metal, can fine-tune their recycling facilities to focus more on these materials.

The synthesis of this literature also highlights future directions for research. The current focus is largely on the recyclable waste stream; however, components such as fire extinguishers, gas cylinders, lithium batteries, for example, are highly relevant, but largely absent from the current literature. In 2024, the annual report from the recycling industry fire protection firm found that there were 373 fires at MRFs and transfer stations across all recycling facility types, including CDW (Staub, 2024). The number of MRF fires is becoming increasingly more common. In the US and Canada, the number of major fires had increased by more than one third since 2017, numbering 390 in 2022. It is estimated that this figure is likely to be much higher as smaller fires often go unreported. These fires are due to several factors including the growing number of plants to deal with demand, major new hazards in the waste streams, and global shifts in policy and attitudes towards CDW management. Finally, a hotter and dryer climate will make it easier for fires to start and spread, destroying millions of dollars in infrastructure (Nugent, 2023).

Despite the extensive information available on CDW waste and its recycling potential, there remain gaps, especially regarding practical applications and stakeholder involvement. Much of the research has focused on measuring material types and volumes, often ignoring the social and economic factors that shape recycling practices. There is a deficit of comprehensive studies that assess how consumer knowledge and habits impact recycling outcomes. Future research could look at pile composition, as it is delivered to the MRF, to feed back to customers in real-time. This is essential for developing effective strategies to divert CDW waste, enhance recovery rates, reduce costs and improve overall environmental results.

The wider implications of these findings suggest that solving the identified issues could improve CDW waste management, enhancing material recovery efficiency and supporting sustainable practices that align with current environmental objectives. Applying advanced CV and DL technologies could automate complex sorting tasks, cut operational costs, and improve the quality of recovered materials, contributing to the circular economy. Successful use of these technologies can also lead to a better understanding of waste composition, guiding future regulations and aligning industry practices with sustainability standards.

The costs involved in generating large-scale datasets are not insignificant (ul Hassan et al., 2022). The literature demonstrates that a wide range of sensors and resolutions have been used previously (see Table 1). However, maintenance schedules are usually not listed. For example, cameras installed in the roof of MRFs may require special access equipment, such as a scissor lift, and trained staff for access and cleaning. Access may be difficult during business hours if there are technical issues. The video quality may degrade during the week if weekend cleaning schedules, during downtime, are adopted. This may affect the outcome or efficiency of the DL models. Furthermore, there would need to be a clear cost benefit for adopting these models as part of the deployment of new digital technologies for assessing and processing CDW.

To tackle these challenges, several areas of exploration are recommended, such as hybrid models that combine different data sources, like sensor data and CV results, to improve classification accuracy. Furthermore, collaborative frameworks involving industry, academia, and tech developers can drive innovation by creating feedback loops that enhance machine learning algorithms based on real-world experiences. There is also a need for studies on the ethical implications of AI in sorting tasks, as this technology is adopted, to ensure that automation does not reduce human oversight and accountability (Joseph et al., 2024).

In conclusion, the literature review gives a thorough overview of the critical issues facing CV and DL in analyzing mixed CDW in MRFs. By detailing both the progress made and the challenges that still exist, it underscores the pressing need for concerted research to realize the full potential of these technologies in achieving sustainable waste management practices.

**Table 1** – Recent studies (since 2019) for C&DW object detection and segmentation assessment.

| Reference | Year | CDW Materials | Camera/Sensor Specification | Method/Model(s) | Setting | Image Resolution | Dataset Size |
|---|---|---|---|---|---|---|---|
| (Prasad and Arashpour, 2024) | 2024 | concrete, rock, stone, bricks; deformed cardboard, high density polyethylene (HDPE) materials; metals (copper, aluminum, steel, iron); translucent polyethylene (LDPE) materials; waste timber and wood | RGB-D | Object segmentation/ RGB-DL depth fusion strategy | Material recovery facility | 1280x720 | train: 2500, validation 713, test 355 (3568 images) |
| (Sirimewan et al., 2024) | 2024 | concrete and aggregates; wood and timber; hard plastic; soft plastic; steel; cardboard and paper; mixed waste | RGB | Object detection; segmentation/ DuoSeg++ DeepLab3+ Unet | Material recovery facility | Non-disclosed | train: 75%, validation 15%, test: 10% (430 images) |
| (Demetriou et al., 2024) | 2024 | concrete; brick; tile; foam; general waste; plaster board; pipes; plastic; wood; stone | RGB | Object detection; segmentation/ YOLOv8 | Conveyor belt | 1920x1200 | train: 1984, test: 573, validation: 570 (3127 images) |
| (Kronenwett et al., 2024) | 2024 | brick; sand-lime brick | Line-Scan Camera | Object detection/ SSD,YOLOv3 Faster R-CNN | Conveyor belt | 300x300 | train: 5000, test: 500, validation: 500 |
| (Demetriou et al., 2023) | 2023 | concrete; brick; tile | RGB | Object detection/ SSD,YOLO Faster R-CNN | Conveyor belt | 1920x1200 | train: 4230, test 1727 (4230 images) |
| (Wang et al., 2023) | 2023 | rebar; bricks; PVC pipes, plastic wires; debris. | RGB | Object segmentation/ Swin Transformer, Twins Transformer, K-Net | Construction site | 620x770 (average) | train: 1696, test: 200, validation 23. (1919 images - different data sources, synthetic images) |
| (Lux et al., 2023) | 2023 | concrete grains; natural stones; ceramics; bituminous grains; glass and others | RGB | Object classification; segmentation; mass regression/ RACNET | Conveyor belt | 8192 × 4096 (database 2) | train: 8000, test: 800, validation: 800 (images – from 2 datasets) |
| (Lin, Kunsen et al., 2022) | 2022 | concrete; brick; stone; ceramic tile; glass; metal scrap; gypsum board; wood; plastic and paper | RGB | Object classification/ ResNet based | Construction site | Non-disclosed (scaled to 224 x224) | train: 36711, validation: 4080, test: 10203 (2836 original images -web crawling/Image augmentation) |
| (Bashkirova et al., 2021) | 2022 | cardboard; soft plastic; rigid plastic; metal | RGB | Object detection; segmentation/ RetinaNet, Mask-RCNN, TridentNet | Conveyor belt | 1920 x 1080 | train: 3002, test: 929, validation 572, unlabeled: 6212 (10715 images) |
| (Zhou et al., 2022) | 2022 | brick; wood; stone; plastic | RGB | Object detection/ Faster-RCNN, YOLO | Construction site | Non-disclosed | train: 80%, test: 10%, validation 10% (3046 images) |
| (Na et al., 2022) | 2022 | concrete; brick; lumber; board; mixed waste | RGB | Object detection; segmentation/ YOLACT | Construction site | 512x512 | (500 images construction site, 288 web crawling) |
| (Sunwoo et al., 2022) | 2022 | concrete; brick; board; mixed waste; wood | RGB | Object classification; detection; segmentation/ YOLACT | Construction site | 512x512 | (599 images construction site, web crawling) |
| (Dong et al., 2022) | 2022 | rock; gravel; earth; packaging; wood; other non-inert, and mixed | RGB | Object segmentation/ Boundary aware Transformer | Weigh bridge top-down view | 1980x1080 | (5366 images) |
| (Li et al., 2022) | 2022 | concrete; machine made brick; fired brick; wood; plaster; plastic; ceramic; carton | RGB-Depth | Object classification; detection/ Mask R-CNN based fusion models | Conveyor belt | Non-disclosed (scaled to 640x640) | train: 70%, test: 30% (3367 images) |
| (Ko et al., 2022) | 2022 | rebar; bricks; PVC pipes; wires; cementitious debris | RGB | Object segmentation/ Detectron2, YOLACT and MMDetection | Construction site | Non-disclosed | (858 images) |
| (Lu et al., 2022) | 2022 | rock; gravel; earth; packaging; wood; other non-inert, and mixed | RGB | Object segmentation/ DeepLabv3+ | Weigh bridge top-down view | 1980x1080 | train: 3515, validation: 754, test: 753 (5366 images) |
| (Chen et al., 2022) | 2022 | cotton gloves; wood blocks; small ferrous; plastic pipe, bamboo; corrugated paper; rebar | RGB-Depth; 3D LiDAR. | Object detection; segmentation/ Mask R-CNN | Construction site | 640x480 | Train: 454, validation: 151, test: 151 (756 images) |
| (Song et al., 2022) | 2022 | brick; woods; plastics; concretes; foams | RGB | Object classification/ | Conveyor belt | Non-disclosed | 125 pictures of each class. train: 100 |
| | | | | VGG-16, ResNet-50, and Transformer | | | samples, test: 25 samples of each type |
| (Lin, K. et al., 2022) | 2022 | concrete; brick; stone; ceramic tile; glass; metal scrap; gypsum board; wood; plastic and paper | RGB | Object classification/ VGG based models | Construction site | Non-disclosed (scaled to 224 x224) | train: 36711, validation: 4080, test: 10203 (2836 original images- web crawling/Image augmentation) |
| (Davis et al., 2021) | 2021 | Second fix timbers; shuttering/formwork timbers; shuttering/formwork ply and particle boards; hard plastics; soft plastics; brick; concrete; cardboard; polystyrene | RGB | Object classification/ VGG based models | Skip bin top-down view | 3000×2250 | (525 images; 84 images empty skip bins) |
| (Chen et al., 2021) | 2021 | Inert (concrete and bricks) and non-inert (wood, plastic and bamboo) | RGB | Percentage of inert waste exceeds certain level (e.g. 50%)/ DenseNet169 | Weigh bridge top-down view | Non-disclosed | train: 70%, validation 15%, test: 15% (1127 records: images, physical properties such as net weight, weight depth) |
| (Lau Hiu Hoong et al., 2020) | 2020 | concrete grains; white stones; grey stones; light colored stones; slate; clay bricks; ceramic tiles; bituminous grains; | RGB | Object classification/ ResNet | Laboratory setting | 6000x4000 (scaled to 256x256) | train: 2000, test: 500 in each of 9 subclasses (36000 labelled database images) |
| | | glass; wood; plastics; steel; paper and cardboard; other | | | | | |
| (Xiao et al., 2020) | 2020 | wood; rubber; brick; concrete | NIR hyperspectral | Object classification/ CNN (class not defined) | Conveyor belt | 100x100/ wavelength 900-1700 nm. | (750 samples) |
| (Wang et al., 2019) | 2020 | Concrete; bricks; plastic bottles; rubber; wood | Hyperspectral camera, laser beam, 3D camera | Object detection/ RCNN Autoencoder | Conveyor belt | 160x160 | train:test 4:1 ratio, (2500 grasping rectangles) |
| (Ku et al., 2020) | 2020 | Bricks; concrete; plastic; metal; wood; rubber | NIR hyperspectral/ 3D camera | Object detection | Conveyor belt | 640x640 | train: 75%, test 25% (2500 samples) |
| (Xiao, Wen et al., 2019) | 2019 | foam; plastic; brick; concrete; wood | RGB/NIR hyperspectral | Object classification/ Single hidden-layer forward neural network | Conveyor belt | 640x480/ wavelength 900-1700 nm. | train: 250, test 150 (samples) |
| (Xiao, W. et al., 2019) | 2019 | Wood; plastic; bricks; concrete; rubber; black bricks | NIR hyperspectral | Object classification | Conveyor belt | wavelength 900-1700 nm. | train: 150 samples each type, test: 166 pieces of woods, 130 pieces of plastics, 142 pieces of red bricks, 150 pieces of concretes, 198 pieces of rubbers, and 158 pieces of black bricks |

**Acknowledgements**
This research was supported by an Australian Research Training Program (HDR) Scholarship.

**Conflict of interest**
The authors declare the following financial interests/personal relationships which may be considered as potential competing interests:

Mostafa Rahimi Azghadi reports financial support was provided by Bingo Industries to assist PhD student research into deep learning in construction and demolition waste management. Matthew Lonergan acts in a consulting and advisory role to Bingo Industries. The remaining authors declare that they have no known competing financial or personal relationships that could have appeared to influence the work reported in this paper.

**Data availability statement**
Data sharing does not apply to this article as no data sets were generated or analyzed during the current study.

**ORCID**
Adrian Langley https://orcid.org/0000-0001-8640-0635
Matthew Lonergan https://orcid.org/0009-0000-3112-395X
Tao Huang https://orcid.org/0000-0002-8098-8906
Mostafa Rahimi Azghadi https://orcid.org/0000-001-7975-3985